\documentclass[runningheads]{llncs}
\usepackage{url}
\usepackage{algorithm}
\usepackage{algpseudocode}
\usepackage{graphicx}
\usepackage{multirow}
\usepackage{amsmath}
\usepackage{tabularx}
\usepackage{caption}
\usepackage{subcaption}
\usepackage{amssymb}

\usepackage{xcolor}

\begin{document}
\mainmatter              
\title{An Isolation Forest Learning Based Outlier Detection Approach for Effectively Classifying Cyber Anomalies}

\titlerunning{An Isolation Forest Learning Based Outlier Detection}
\author{Rony Chowdhury Ripan$^1$ \and Iqbal H. Sarker$^{1,*}$ \and Md Musfique Anwar$^2$ \and Md. Hasan Furhad$^3$ \and Fazle Rahat$^1$ \and Mohammed Moshiul Hoque$^1$ \and Muhammad Sarfraz$^4$}

%
\authorrunning{Ripan et al.}

\institute{$^1$ Dept of Computer Science \& Engineering, Chittagong University of Engineering \& Technology, Chittagong-4349, Bangladesh.\\
$^2$Jahangirnagar University, Dhaka, Bangladesh.\\
$^3$ Canberra Institute of Technology, Canberra, Australia. \\
$^4$ Department of Information Science, Kuwait University, Shadadiya, Kuwait.
$*$Correspondence: iqbal@cuet.ac.bd
}
\maketitle 

\begin{abstract}
Cybersecurity has recently gained considerable interest in today's security issues because of the popularity of the Internet-of-Things (IoT), the considerable growth of mobile networks, and many related apps. Therefore, detecting numerous cyber-attacks in a network and creating an effective intrusion detection system plays a vital role in today's security. 
However, it is difficult to accurately model cyber threats since modern security databases contain large number of security features that could include \textit{Outliers}. In this paper, we present an \textit{Isolation Forest Learning-Based Outlier Detection Model} for effectively classifying cyber anomalies. In order to evaluate the efficacy of the resulting Outlier Detection model, we also use several conventional machine learning approaches, such as Logistic Regression (LR), Support Vector Machine (SVM), AdaBoost Classifier (ABC), Naive Bayes (NB), and K-Nearest Neighbor (KNN). The effectiveness of our propsoed Outlier Detection model is evaluated by conducting experiments on Network Intrusion Dataset with evaluation metrics such as precision, recall, F1-score, and accuracy. Experimental results show that the classification accuracy of cyber anomalies has been improved after removing outliers.

\end{abstract}
\section{Introduction}
\label{intro}

The modern digital world is overwhelmed by an unprecedented amount of data in various domains \cite{sarker2020mobile}. Network intrusion is related to cybersecurity data \cite{sarker2020cybersecurity}, which refers to any attempt to undermine the host and network's confidentiality, credibility, or availability \cite{buczak2015survey}, and is one of cyberspace's most common threats. An intrusion detection system (IDS) is a network security device that tracks real-time network traffic and can warn or take corrective action when unauthorized transmissions are discovered \cite{samrin2017review}. The fundamental motive for detecting the intrusion is to develop classifier accuracy that can effectively recognize the invasive behavior \cite{yin2017deep}. Unfortunately, new security threats are quickly emerging since the attackers become more advanced than before and as a result, the possibility of damaging vital infrastructures increases dramatically in recent time. Thus, it is very important for IDS to detect and deal with novel attacks.

Supervised classification techniques can learn well from a
security dataset that can play a major role in IDS  \cite{sarker2020cybersecurity,seufert2007machine}. However, modeling cyber-attacks effectively is problematic due to the high dimensions of security features and the presence of outliers in today's security datasets \cite{sarker2020cybersecurity}.
Outliers are the data that differs significantly from normal behavior, also known as noisy instances in data \cite{sarker2019machine} \cite{sarker2020cybersecurity}. Outliers found in a security model may also cause various problems, such as over-fitting problems in a classification model. The computing time and cost can be very high to train such model due to the scarcity of model generalization. As a result, the performance of security model  degrades in terms of classification accuracy.

In this paper, we propose a model based on the Isolation Forest algorithm. At first, we perform necessary preprocessing steps like categorical feature encoding, feature scaling \cite{sarker2020context} to extract fifteen essential features to fit into the proposed model. Finally, we applied five popular classification algorithms \cite{sarker2019effectiveness} such as Logistic Regression (LR), Support Vector Machine (SVM), AdaBoost Classifier (ABC), Naive Bayes (NB), and K-Nearest Neighbor (KNN) to evaluate the performance of our system. 

The rest of the paper is organized as follows: Section \ref{sec:2} reviews related works of the outlier detection system and network intrusion detection system. In Section \ref{sec:3}, we present our proposed outlier detection model. 
Our experimental design is described in Section \ref{sec:4}, including a brief discussion. Finally, Section \ref{sec:5} concludes the paper and underlines the future work.

\section{Related Work}
\label{sec:2}

A considerable amount of research has been devoted to the problem of outlier detection. For example, Mascaro et al. \cite{mascaro2014anomaly} proposed an anomaly detection system by combining dynamic and static Bayesian network models to improve anomaly detection performance. The authors in \cite{mohamed2011outlier} proposed a method using standard SVM to classify the sensor node data into different categories, such as a local outlier or cluster outlier, or network outlier. Sun et al. \cite{sun2016detecting} presented an anomalous user behavior detection framework that applies an extended version of the Isolation Forest algorithm to an enterprise dataset to isolate anomalous instances. Stripling et al. \cite{stripling2018isolation} represented the iForestCAD approach that computes conditional anomaly scores to detect fraud. Ben-Gal et al. \cite{ben2005outlier} presented multiple approaches for outlier identification that can be divided between univariate vs. multivariate techniques as well as parametric vs. nonparametric procedures. 
Aggarwal et al. \cite{aggarwal2001outlier} proposed a method which is capable of finding lower-dimensional projections that are locally sparse. As a result, their method is suitable for outlier detection in high dimensional data. 


There are numerous studies on network intrusion detection. For Instance, Song et al. \cite{song2020vehicle} represented a deep convolutional neural network based intrusion detection system. Shapoorifard et al. \cite{shapoorifard2017intrusion} proposed a Hybrid IDS model to improve the KNN classifier in the remaining intrusion detection task, which combines K-MEANS clustering and KNN classification. Yan et al. \cite{yan2010new} proposed a new network intrusion detection method that builds an improved transductive SVM by introducing simulated annealing to degenerate the optimization problem and then apply Support Vector Classifier. However, all these foregoing approaches did not consider the impact of outlier instances on the classification results. Our proposed framework considers this factor and can improve the classification accuracy of cyber anomalies in a network intrusion dataset.  


\section{Methodology}
\label{sec:3}

\begin{figure}
    \centering
    \includegraphics[width = 12cm]{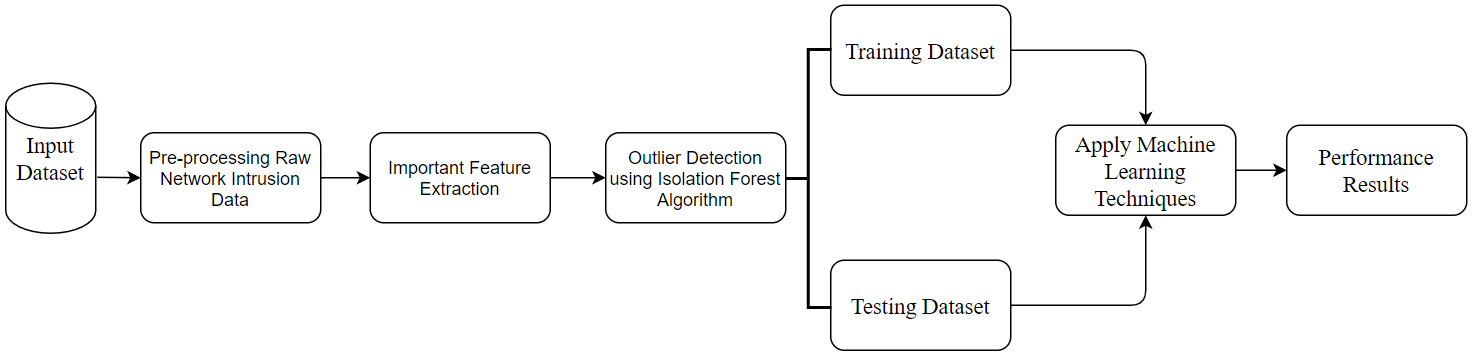}
    \caption{The Proposed Methodology}
    \label{fig:1}
\end{figure}

In order to provide complete coverage of the system, we model an approach that has different stages, as presented in Fig. \ref{fig:1}. Firstly, the pre-processing is performed on raw data to normalize the values of the features. Secondly, we select important features using Recursive Feature Elimination (RFE) \cite{sarker2020intrudtree}. Next, we apply the Isolation forest algorithm to train the instances of the dataset. Finally, we analyze the effectiveness of the proposed outlier detection model.

\subsection{Dataset \& Data Preprocessing}
To develop an efficient outlier detection model, it is necessary to understand the characteristics of network data. In this study, we used a network intrusion dataset that is publicly available in Kaggle \cite{bhosale_2018}. This dataset contains a total of 41 features of which, 3 of those features are \textit{Nominal}, 15 of those features are \textit{Float}, and the remaining 23 features are \textit{Integer}. The target feature set consists of two (normal and anomaly) types of class variables. The dataset was created by simulating a standard local area network of the US Air Force, which is exposed to numerous cyber attacks known as intrusions \cite{bhosale_2018}. In terms of data distribution, all the features are not identical and differ one from another. So, the pre-processing of this raw data is needed to build the proposed Isolation Forest Learning-based outlier detection model.
Data pre-processing requires both encodings and scaling on the features based on the specified network intrusion dataset's characteristics. Although most of the features in this intrusion dataset are quantitative, it has 3 qualitative features such as: \textit{protocol\_type}, \textit{flag}, and \textit{service}. These three features need to be encoded to build the Isolation Forest Learning-based outlier detection model. In this study, we used ``Label Encoding" instead of ``One Hot Encoding" to avoid the feature increase problem of ``One Hot Encoding" \cite{sarker2020intrudtree}. As we described earlier, the security features' values vary from feature to feature in different ranges. So, We used a Standard Scaler to normalize the safety features with an average value of 0 and a standard deviation of 1.

\subsection{Important Features Extraction} 
This network intrusion dataset has 41 features that are used in our modeling purpose. Having too many irrelevant features can increase overfitting, training time and decrease the accuracy. Therefore, selecting the essential features that contribute most to the prediction variable or output in a specific dataset is necessary \cite{sarker2020intrudtree}. In this work, we apply the Recursive Feature Elimination (RFE) method for this purpose. RFE is a wrapper method with two important configuration options: the choice in the number of features to select and the choice of the algorithm that helps to choose those features. We choose 15 features that have been selected using Extra Tree Classifier with the help of the RFE technique.

\subsection{Outlier Detection with Isolation Forest}
Isolation forest \cite{liu2008isolation} is an unsupervised outlier detection algorithm that works rather than profiling normal points on the notion of isolating outliers. The most common outlier detection techniques are focused on building a profile of ``normal" instances, after which the outliers are recorded as those that do not adhere to the normal profile in the dataset. Isolation Forest explicitly isolates outlier instances in the dataset. Two quantitative properties of outlier data points in a sample are the basis for Isolation Forest. They are fewer and have feature values that are very different from those of ``normal" instances. They are the minority.

\begin{figure}
    \centering
    \includegraphics[width = 12cm]{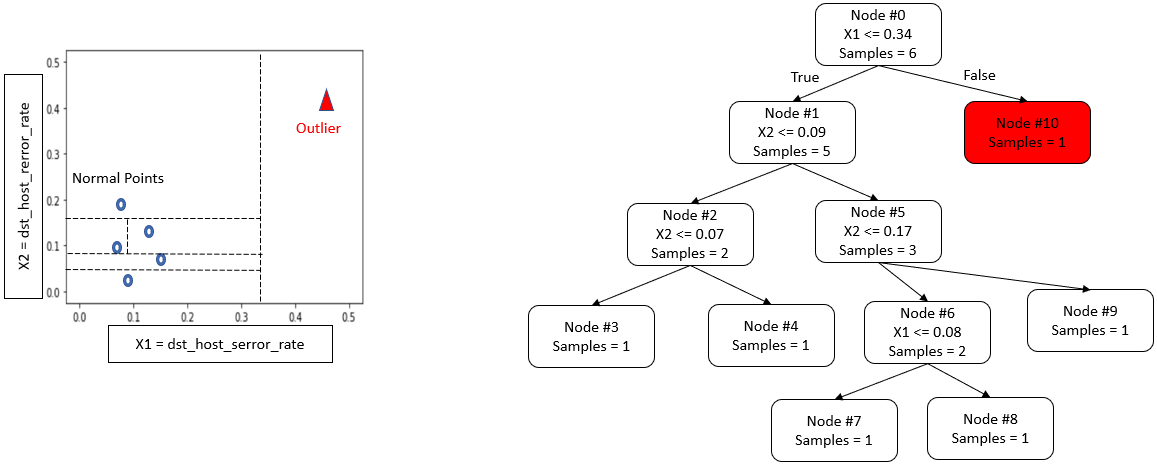}
    \caption{Example of an isolation tree (iTree) on Network Intrusion data. The red triangle represents outlier data point}
    \label{fig:2}
\end{figure}

Isolation Forest builds an ensemble of ``iTrees" (Isolation Trees) on network intrusion dataset similar to the example in Fig. \ref{fig:2}. The algorithm takes \textit{n} random samples of size \textit{m} from a given dataset. For each random sample, the ``iTree" is built by splitting the sub-sample instances over a split value of a randomly selected feature so that the instances whose corresponding feature value is smaller than the split value go left, and the others go right, and the process continues recursively until the tree is entirely built. The split value is selected at random between the minimum and maximum values of the selected feature. In addition, the ``iTree" comprises two groups of nodes; internal nodes and external nodes. Internal nodes are non-leaf, containing the split value, split features, and pointing to two child subtrees. External nodes are leaf nodes that can not be further separated and reside at the bottom of the tree and carry the unbuilt subtree's size to measure the outlier value. Finally, outliers are the points that have shorter average path lengths on the ``iTrees."

The outlier score of an instance is calculated based on the observation that the structure of ``iTrees" is similar to that of Binary Search Trees (BST): the termination of the external node ``iTree" corresponds to the failure of the BST search. Consequently, the estimate of the average \textit{h(x)} for the termination of the external node is the same as an unsuccessful search in BST. That is,

\begin{equation}
    {\displaystyle c(m)={\begin{cases}2H(m-1)-{\frac {2(m-1)}{m}}&{\text{for }}m>2\\1&{\text{for }}m=2\\0&{\text{otherwise}}\end{cases}}}
    \label{eqn:1}
\end{equation}

where $n$ is the testing data size, $m$ is the size of the sample set, and $H$ is the harmonic number, which can be estimated by

\begin{equation}
    {\displaystyle H(i)=ln(i)+ 0.5772156649 \text{ (Euler-Mascheroni constant.)}}
\end{equation}

The value of $c(m)$ in Equation \ref{eqn:1} is the average $h(x)$ for a given $m$. Then it is used to normalize $h(x)$ to get an estimate of outlier score for a given instance $x$ as:

\begin{equation}
    {\displaystyle s(x,m)=2^{\frac {-E(h(x))}{c(m)}}}
\end{equation}

where $E(h(x))$ is the average value of $h(x)$ from a collection of ``iTrees". Finally, instance $x$ is assigned to outlier if the value of $s$ is close to 1, otherwise \textit{x} is likely to be considered normal instance if the value of $s$ is smaller than 0.5. After detecting outliers using the Isolation Forest algorithm, we remove all the outliers to evaluate the system's improved performance in terms of classification accuracy. To achieve this goal, we employ five popular machine learning classification techniques \cite{sarker2019effectiveness}, such as Logistic Regression (LR), Support Vector Machine (SVM), AdaBoost Classifier (ABC), Naive Bayes (NB), and K-Nearest Neighbor (KNN) to classify cyber anomalies in the Network Intrusion dataset.

\section{Experimental Evaluation}
\label{sec:4}
In this section, we describe all the performance metrics used in this work to test our proposed model. In order to test our model, we use Precision, Recall, Accuracy, and F1-score.

\begin{eqnarray}
   Precision = \frac{TP}{TP + FP}\\
   Recall = \frac{TP}{TP + FN}\\
   Accuracy = \frac{TP + TN}{TP + TN + FP + FN}\\
   F1-score = 2 \times \frac{Precision \times Recall}{Precision + Recall}
\end{eqnarray}

\subsection{Experimental Setup}
All tests are performed on an 8 GB RAM Intel Core i5 2.50GHz CPU. The proposed Isolation Forest Learning-Based Outlier Detection model is implemented in Python with packages scikit-learn under OS Windows 10.

\subsection{Implementation \& Performance Evaluation}

\begin{figure}
     \centering
     \begin{subfigure}[b]{0.3\textwidth}
         \centering
         \includegraphics[height = 6cm, width=6 cm]{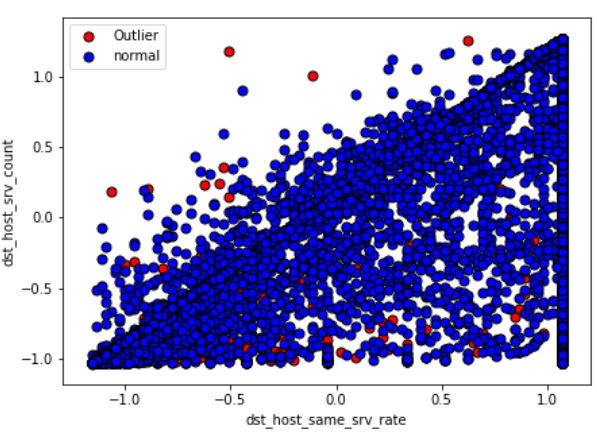}
         \caption{Original Dataset}
         \label{fig:3.1}
     \end{subfigure}
     \hfill
     \begin{subfigure}[b]{0.5\textwidth}
         \centering
         \includegraphics[height = 6cm, width=6 cm]{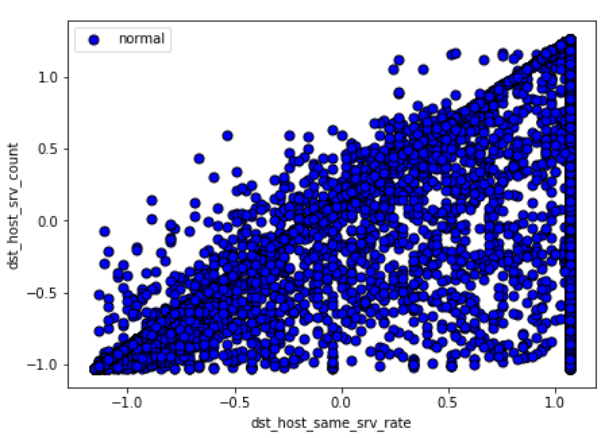}
         \caption{Without Outlier}
         \label{fig:3.2}
     \end{subfigure}
     \hfill
       \caption{Scatter plot of Network Intrusion Data}
        \label{fig:3}
\end{figure}

As we stated earlier in Section \ref{sec:3}, first we preprocess raw dataset for fitting into the Isolation Forest model. At first, all the categorical data are transformed into numerical value using Label Encoder. Next, all the features are normalized using Standard Scaler. Then, Recursive Feature Elimination (RFE) is used for feature selection. In this work, we select 15 feature 
out of 41 features. Next, data is fitted into the Isolation Forest Model that assigns a value of 1 to an instance if it is a normal instance, otherwise assigns -1 if it is an outlier. A scatter plot of normal and outlier data points of network intrusion data is represented in Fig. \ref{fig:3.1}. 

\begin{figure}
    \centering
    \includegraphics[height = 7cm]{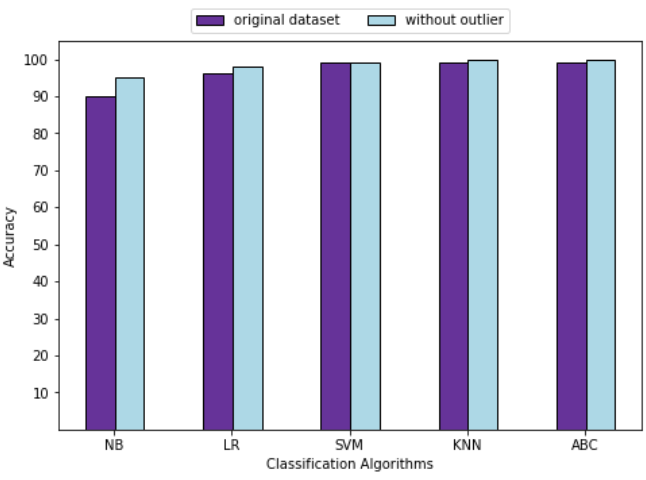}
    \caption{Bar Chart of Classification Accuracy}
    \label{fig:4}
\end{figure}

After that, all the outliers have been removed to improve the classification accuracy of cyber anomalies. A scatter plot of network intrusion data after removing outliers is represented in Fig. \ref{fig:3.2}. For measuring classification accuracy, we apply five popular classification algorithms; Logistic Regression (LR), Support Vector Machine (SVM), AdaBoost Classifier (ABC), Naive Bayes (NB), and K-Nearest Neighbor (KNN). We observe that the classification accuracy of cyber anomalies has been improved after removing outliers except for SVM. The proposed model also demonstrates improved performance in terms of Precision, Recall, F1-score values. Table \ref{tab:3} shows that the classifiers KNN, SVM, NB, LR and ABC have accuracy of 99\%, 99\%, 90\%, 96\% and 99\%, respectively. After removing outliers, KNN, SVM, NB, LR, and ABC classifiers have an accuracy of 100\%, 99\%, 95\%, 98\%, and 100\%, respectively. For a better view, a graphical representation of accuracy comparison has shown in Fig \ref{fig:4}.

\begin{figure}
     \centering
     \begin{subfigure}[b]{0.3\textwidth}
         \centering
         \includegraphics[height = 6cm, width=6 cm]{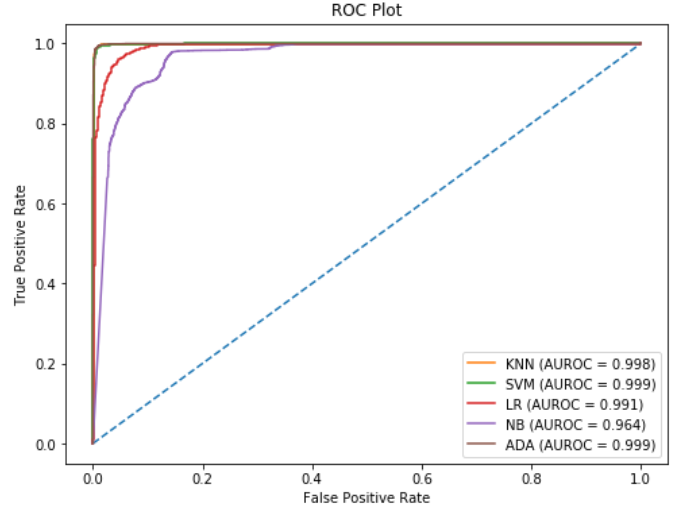}
         \caption{Original Dataset}
         \label{fig:5.1}
     \end{subfigure}
     \hfill
     \begin{subfigure}[b]{0.5\textwidth}
         \centering
         \includegraphics[height = 6cm, width=6 cm]{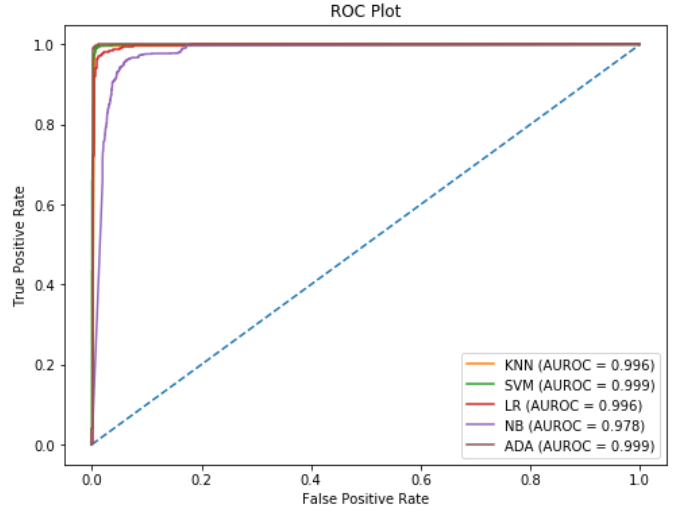}
         \caption{Without Outlier}
         \label{fig:5.2}
     \end{subfigure}
     \hfill
       \caption{ROC curve of various machine learning classification models}
        \label{fig:5}
\end{figure}

\begin{table}
    \centering
    \caption{comparison of Precision, Recall, Accuracy, F1-score}
    \label{tab:3}
    \vspace{0.2cm}
    \begin{tabular}{|c|c|c|c|c|c| } 
        \hline
        Dataset & classification & Accuracy & Precision & Recall & F1-score\\
        & models & (\%) & (\%) & (\%) & (\%)\\
        \hline
        \multirow{5}{5em}{Original dataset}& KNN & 99 & 99 & 99 & 99\\
        & SVM & 99 & 99 & 99 & 99\\
        & NB & 90 & 90 & 90 & 90\\
        & LR & 96 & 96 & 96 & 96\\
        & ABC & 99 & 99 & 99 & 99\\
        \hline
        \multirow{5}{5em}{Without Outlier}& KNN & 100 & 100 & 100 & 100\\
        & SVM & 99 & 99 & 99 & 99\\
        & NB & 95 & 95 & 94 & 95\\
        & LR & 98 & 98 & 98 & 98\\
        & ABC & 100 & 100 
        & 100 & 100\\
        \hline
    \end{tabular}
    
\end{table}

Moreover, to evaluate the performance of our Isolation Forest learning-based outlier detection model, the Receiver Operating Characteristic (ROC) curve of the five classifiers for the dataset with and without outliers have been shown in Fig. \ref{fig:5.1} and Fig. \ref{fig:5.2}. From these two figures, it is observed that after removing outliers, LR and NB have a better area under the ROC curve (AUC) of 0.996 and 0.997, respectively, than the original dataset.

\section{Conclusion}
\label{sec:5}
In this paper, we present an Isolation Forest Learning-Based outlier detection approach to classify Cyber anomalies effectively. In our approach, we have first considered feature selection according to their importance and then applied the Isolation Forest to detect outliers. After that, we remove all the outliers to make the network intrusion detection model more effective regarding classification accuracy. Finally, by performing several tests on the Network Intrusion Dataset, the efficacy of the proposed outlier detection model is tested. Experimental results show that the classification accuracy of cyber anomalies has been improved after the removal of outliers.
For future work, the efficacy of our propsoed Isolation Forest Learning-based Outlier Detection model can be evaluated by gathering large data sets with more security feature dimensions in IoT security services and also evaluate their efficiency at the application level in the cybersecurity domain.

%
%
\bibliographystyle{splncs04} 
\bibliography{anomaly} 

\end{document}